\documentclass[runningheads]{llncs}

\usepackage[T1]{fontenc}
\usepackage{amssymb}
\usepackage{graphicx}
\usepackage{caption}
\usepackage{booktabs}
\usepackage{amsmath}
\usepackage{svg}
\usepackage{hyperref}
\usepackage{makecell}
\usepackage{graphicx,verbatim}

\usepackage{tabularx}
\usepackage{booktabs}
\begin{document}
\title{Pretraining of Medical Visual Encoders Toward Multi-modal Large Language Models}
%
\begin{comment}  %% Removed for anonymized MICCAI 2025 submission
\author{First Author\inst{1}\orcidID{0000-1111-2222-3333} \and
Second Author\inst{2,3}\orcidID{1111-2222-3333-4444} \and
Third Author\inst{3}\orcidID{22fi22--3333-4444-5555}}
%
\authorrunning{F. Author et al.}
\institute{Princeton University, Princeton NJ 08544, USA \and
Springer Heidelberg, Tiergartenstr. 17, 69121 Heidelberg, Germany
\email{lncs@springer.com}\\
\url{http://www.springer.com/gp/computer-science/lncs} \and
ABC Institute, Rupert-Karls-University Heidelberg, Heidelberg, Germany\\
\email{\{abc,lncs\}@uni-heidelberg.de}}

\end{comment}

\author{Tianyou Jiang}  %% Added for anonymized MICCAI 2025 submission
\authorrunning{Tianyou Jiang et al.}
\institute{University of Bern, Bern, Switzerland \\
    \email{tianyou.jiang.research@gmail.com}}

\maketitle              % typeset the header of the contribution
\begin{abstract}
Multimodal Large Language Models (MLLMs) commonly reuse visual encoders pretrained with CLIP, although the features of these ViTs are ultimately consumed by autoregressive LLMs. We refer to this mismatch as the semantic-interface gap and introduce MedMLIP, a framework that pretrains the visual encoder through report generation with a frozen LLM, while employing Local Relational Distillation (LRD) to preserve relationships among visual patches to avoid visual collapse. We pretrain MedMLIP on IU-Xray and Open-PMC-300K and evaluate the resulting encoders on VQA-RAD and SLAKE. Only the ViT is transferred, while the guiding LLM and projector are replaced, allowing us to assess cross-LLM transferability. Our cross-LLM transfer experiments demonstrate the value of pretraining visual encoders for their autoregressive LLM interface while trying to preserve more fine-grained visual information. Code and the pretrained model are available at \url{https://github.com/SkyCol/MedMLIP}.

\keywords{Medical VLMs \and MLLMs \and LLMs}
% Authors must provide keywords and are not allowed to remove this Keyword section.

\end{abstract}
\section{Introduction}

In a multimodal large language model (MLLM), the visual encoder is the component that directly observes the image. Yet, in most existing MLLMs, this encoder is pretrained for an interface fundamentally different from the autoregressive interface through which it is ultimately used for Visual Question Answering (VQA). Visual encoders such as CLIP~\cite{radford2021learning} are
typically optimized through contrastive vision--language pretraining, where visual
representations are aligned with a paired text encoder. When constructing an
MLLM, however, this text encoder is discarded, and the pretrained visual
features are instead projected into the embedding space of an autoregressive
large language model (LLM). Consequently, the visual representation is learned primarily for cross-modal similarity matching rather than for token-level interpretation and generation by the LLM that eventually consumes it.

We refer to this discrepancy as a \emph{semantic-interface gap}. The problem is not whether the visual encoder contains useful semantic
information, but whether this information is represented in a form that can be
effectively used by an autoregressive LLM. This distinction
becomes particularly important in medical MLLMs, where successful reasoning
requires both high-level semantic understanding and preservation of subtle,
fine-grained visual evidence.

Existing medical MLLMs have already provided empirical evidence that the choice of
visual representation can have a substantial impact on downstream performance.
For example, LLaVA-Med~\cite{li2023llavamed} reports performance differences
across visual initialization strategies when training its medical MLLM.
Under comparable training settings, replacing the general-domain CLIP
initialization with the BioMedCLIP~\cite{zhang2023biomedclip} visual encoder
increases the average score across VQA-RAD, SLAKE, and PathVQA from 73.90 to
75.40. The BioMedCLIP initialization notably improves SLAKE open-ended
performance from 82.30 to 87.11 and PathVQA open-ended performance from 37.59
to 39.60. These results suggest that improving the visual representation can enhance the
capability of the resulting MLLM even without changing the downstream datasets
or the fundamental multimodal modeling paradigm.

However, simply replacing a general-domain visual encoder with a
domain-specific contrastive encoder does not fully address the interface
mismatch. BioMedCLIP and related models improve biomedical visual semantics,
but their visual representations are still primarily optimized for alignment
with a paired contrastive text encoder. This motivates a more direct question: \emph{Can we pretrain the visual encoder for its eventual role within an MLLM while preserving the fine-grained visual evidence required for medical tasks?}

To answer this question, we introduce \textbf{MedMLIP}, a Medical Multimodal Language–Image Pretraining framework that explicitly optimizes
the visual encoder for downstream language-model consumption. MedMLIP contains
two complementary components. First, we connect the trainable visual encoder
to a \emph{frozen autoregressive LLM} through a visual projector and pretrain the
encoder using medical report generation. Because the LLM remains frozen, gradients from the generation loss are
propagated through the visual projector to the visual encoder, adapting the
visual pathway toward the representational interface expected by the LLM.
This encourages the ViT to produce visual tokens that can be directly consumed
by an autoregressive language model.

Direct language-level supervision, however, introduces a second problem.
Report generation primarily rewards high-level semantic information and may
suppress local distinctions that are not immediately necessary for predicting
the report. Such degradation is particularly undesirable in medical images,
where small and localized findings can determine the answer to a downstream
question. We therefore introduce \textbf{Local Relational Distillation (LRD)},
which uses a frozen pretrained visual encoder as a structural anchor. Rather than matching individual patch features, LRD preserves the pairwise
relational geometry among visual patches. This allows the student encoder to
adapt to the LLM generation interface while retaining the inter-patch
organization of the original representation.

MedMLIP therefore optimizes two complementary properties of the visual
representation: \emph{compatibility with autoregressive LLMs} through report
generation and \emph{fine-grained evidence preservation} through local
relational distillation. Conceptually, our objective is not to make the visual encoder more language-like
at the expense of visual perception, but to make it
\emph{LLM-ready while preserving fine-grained visual detail}.

We evaluate this hypothesis under two complementary medical data regimes:
the radiology-specific IU-Xray dataset~\cite{demner2016preparing} and
Open-PMC-300K, a larger-scale biomedical image--text subset that we constructed
from Open-PMC~\cite{baghbanzadeh2025advancing}. The former provides
domain-specific chest X-ray--report supervision, whereas the latter contains
300,000 image--text pairs spanning diverse biomedical imaging modalities.
The resulting visual encoders are transferred to MLLMs and evaluated
on VQA-RAD~\cite{lau2018dataset} and SLAKE~\cite{liu2021slake} under the same
training protocol. As summarized in Table~\ref{tab:visual_encoder_transfer}, MedMLIP improves downstream
medical VQA performance, demonstrating that MLLM-oriented visual pretraining
can produce more transferable representations than the widely adopted
contrastive pretraining paradigm.

% Our main contributions are threefold. First, we identify a \emph{semantic-interface gap} between conventional
% contrastive visual pretraining and the autoregressive interface employed by
% modern MLLMs, motivating a visual pretraining paradigm that explicitly
% accounts for how the resulting representations are consumed by an LLM. Second, we introduce \textbf{MedMLIP}, which combines autoregressive
% pretraining through a frozen LLM with \textbf{Local Relational Distillation}
% to learn LLM-compatible representations while preserving fine-grained visual
% evidence. Third, we evaluate MedMLIP using two complementary pretraining regimes,
% the radiology-specific IU-Xray dataset and the diverse biomedical
% Open-PMC-300K subset, and demonstrate consistent improvements after
% transferring the pretrained visual encoders to MLLMs on VQA-RAD and SLAKE. These results show that pretraining visual representations specifically for
% use in MLLMs can improve medical VQA performance without modifying the
% downstream MLLM architecture.

Our main contributions are summarized as follows.

\noindent\textbf{a.}
We identify a \emph{semantic-interface gap} between conventional contrastive
visual pretraining and the autoregressive interface employed by modern MLLMs.
This motivates a visual pretraining paradigm that explicitly accounts for how
visual representations are subsequently consumed by an LLM.

\noindent\textbf{b.}
We introduce MedMLIP, which combines autoregressive pretraining
through a frozen LLM with Local Relational Distillation to learn
LLM-compatible representations while preserving fine-grained visual evidence.

\noindent\textbf{c.}
We evaluate MedMLIP using two complementary pretraining regimes: the
radiology-specific IU-Xray dataset and the diverse biomedical Open-PMC-300K
subset. We transfer the pretrained visual encoders to different MLLMs and
evaluate them on VQA-RAD and SLAKE. The results show that visual
representations pretrained specifically for MLLMs can improve medical VQA
performance without modifying the downstream MLLM architecture. We also
release the ViT pretrained on Open-PMC-300K under the guidance of Qwen3B as a
reusable visual encoder for medical MLLMs. The code and pretrained model are
publicly available at \url{https://github.com/SkyCol/MedMLIP}.

\section{Related Works}

\subsection{Visual Representation Pretraining}
Visual representation learning has progressed from supervised pretraining on
labeled datasets to self-supervised and vision--language pretraining.
Supervised objectives mainly encourage category-level discrimination, whereas
methods such as MAE~\cite{he2022mae} and DINO~\cite{caron2021dino} learn
transferable visual structure through masked reconstruction and self-distillation,
respectively. Vision--language models such as CLIP~\cite{radford2021learning}
and BiomedCLIP~\cite{zhang2023biomedclip} further use large-scale image--text
pairs to associate visual representations with natural-language concepts.
These approaches provide strong initialization for downstream tasks, but their
objectives are intended for classification, reconstruction, or bidirectional
image--text alignment rather than autoregressive multimodal generation.

The choice of visual encoder and its pretraining also affects downstream MLLM
performance. In the medical domain, LLaVA-Med~\cite{li2023llavamed} reports a
modest overall advantage when initialized with BiomedCLIP rather than
general-domain CLIP. This result highlights the value of domain-appropriate
visual representations and motivates pretraining the visual encoder for its
eventual role in an autoregressive MLLM while retaining fine-grained visual
evidence.

\subsection{Visual--Language Alignment in Multimodal Large Language Models}
\begin{figure}[t]
\centering 
\includegraphics[width=1\textwidth, 
                 height=1\textheight,
                 keepaspectratio,     
                 trim=0cm 5cm 0cm 0cm, 
                 clip]{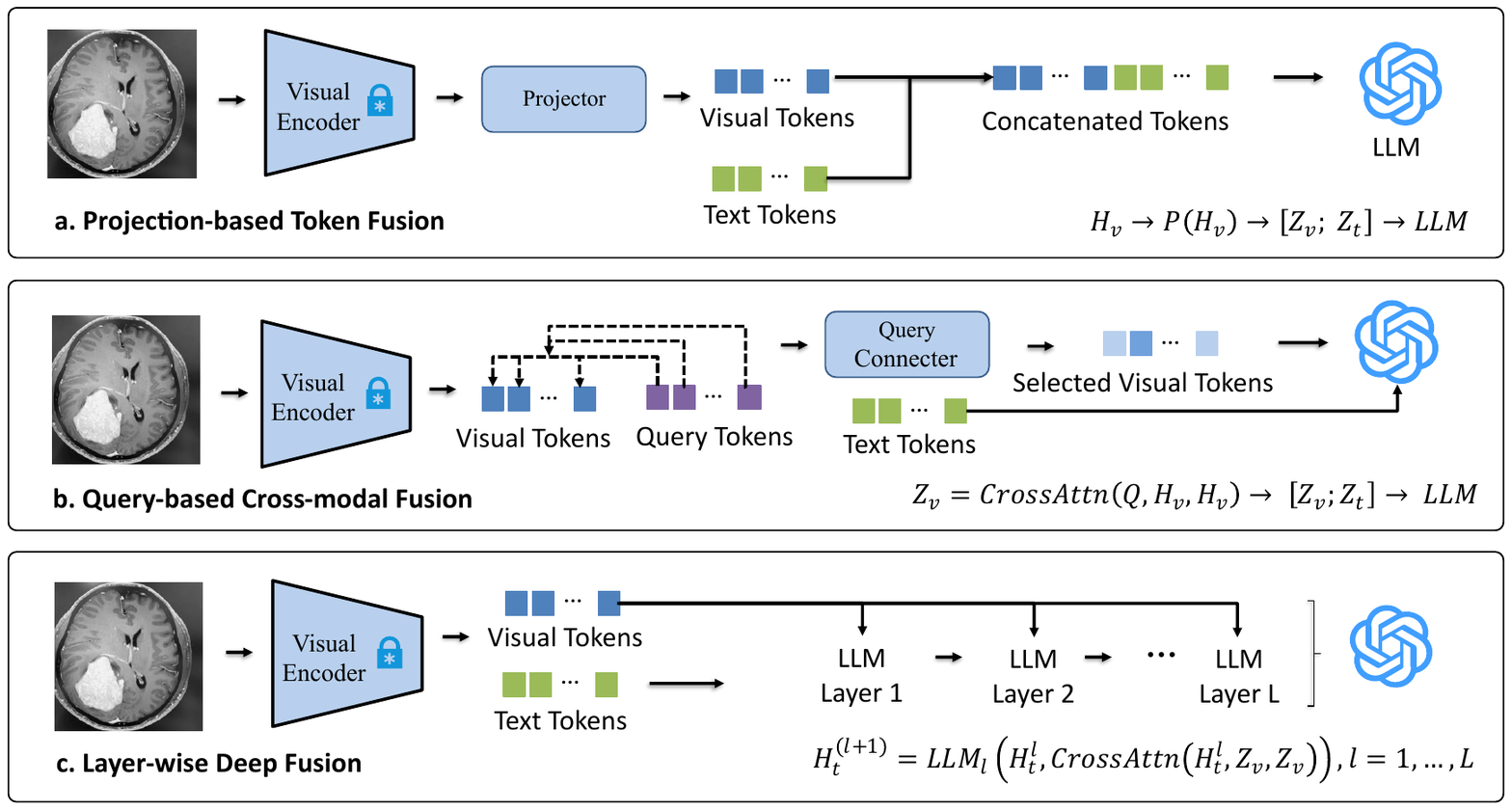}
    \caption{Representative vision--language alignment paradigms in MLLMs: 
    (a) projection-based token fusion, (b) query-based cross-modal fusion, and 
    (c) layer-wise deep fusion. Although differing in how visual information is 
    mapped and integrated, all paradigms fundamentally depend on the quality of 
    the visual representations provided to the LLM.}
    \label{fig:Visual_Language_Alignment}
    \label{fig:Visual_Language_Alignment}
\end{figure}

Existing MLLMs employ different mechanisms to integrate visual representations with language models, which can be broadly grouped into three paradigms, as illustrated in Fig.~\ref{fig:Visual_Language_Alignment}. Projection-based methods, such as LLaVA~\cite{liu2023llava} and Qwen2-VL~\cite{wang2024qwen2vl}, map visual features into the language embedding space through a linear layer, an MLP, or a patch-merging module, and concatenate the resulting visual tokens with text tokens for joint processing by the LLM. Query-based methods, including BLIP-2~\cite{li2023blip2} and Qwen-VL~\cite{bai2023qwenvl}, employ learnable queries and cross-attention to selectively extract and compress information from the visual features before injecting it into the LLM. Layer-wise fusion methods, such as Flamingo~\cite{alayrac2022flamingo} and CogVLM~\cite{wang2023cogvlm}, enable visual--textual interaction within multiple language-model layers through cross-attention or specific visual experts.

Despite architectural differences, these approaches primarily focus on how visual features are mapped, compressed, and injected into the language model. Their effectiveness nevertheless depends fundamentally on the quality of the representations produced by the visual encoder. Many existing MLLMs initialize this component using encoders pretrained with contrastive objectives, which are not explicitly optimized for autoregressive language generation. 

% \subsection{Medical Multimodal Large Language Models}

\section{Methods}
% \subsection{Local Relational Distillation}

% Given an input image \(x\), we use the frozen original CLIP ViT \(E_0\) as the visual teacher and the trainable ViT \(E_\theta\) as the student. After removing the CLS token, their contextualized patch-level visual features are denoted as
% \begin{equation}
% H_0 = E_0^{\mathrm{patch}}(x)
% = \left[ h_0^1, h_0^2, \ldots, h_0^P \right],
% \end{equation}
% \begin{equation}
% H_\theta = E_\theta^{\mathrm{patch}}(x)
% = \left[ h_\theta^1, h_\theta^2, \ldots, h_\theta^P \right],
% \end{equation}
% where \(P\) denotes the number of image patches.

% Each patch feature is first normalized along the feature dimension:
% \begin{equation}
% \bar{h}_0^i =
% \frac{h_0^i}{\left\| h_0^i \right\|_2},
% \qquad
% \bar{h}_\theta^i =
% \frac{h_\theta^i}{\left\| h_\theta^i \right\|_2}.
% \end{equation}

% We then compute the patch-to-patch relational matrices:
% \begin{equation}
% S_0^{ij}
% =
% \left\langle
% \bar{h}_0^i,
% \bar{h}_0^j
% \right\rangle,
% \qquad
% S_\theta^{ij}
% =
% \left\langle
% \bar{h}_\theta^i,
% \bar{h}_\theta^j
% \right\rangle.
% \end{equation}

% The Local Relational Distillation loss is defined as
% \begin{equation}
% \mathcal{L}_{\mathrm{LRD}}
% =
% \frac{1}{P(P-1)}
% \sum_{i \ne j}
% \left(
% S_\theta^{ij}
% -
% \operatorname{sg}
% \left(
% S_0^{ij}
% \right)
% \right)^2,
% \end{equation}
% where \(\operatorname{sg}(\cdot)\) denotes the stop-gradient operation.

% The final training objective is
% \begin{equation}
% \mathcal{L}
% =
% \mathcal{L}_{\mathrm{gen}}
% +
% \lambda
% \mathcal{L}_{\mathrm{LRD}}.
% \end{equation}
\subsection{LLM-guided Pretraining by Auto-Regressive Objective}

Given a medical image \(x\) and its corresponding report \(y=\{y_1,\ldots,y_T\}\), we pretrain the visual encoder through a frozen LLM generation interface:
\begin{equation}
x \rightarrow E_{\theta} \rightarrow P_{\psi} \rightarrow F_{\phi} \rightarrow y .
\end{equation}
Here, \(E_{\theta}\) denotes the ViT visual encoder, \(P_{\psi}\) denotes the visual projector, and \(F_{\phi}\) denotes the frozen LLM.

The image is first encoded into visual features and then projected into the LLM embedding space:
\begin{equation}
H_{\theta}=E_{\theta}(x),
\qquad
V_{\theta,\psi}=P_{\psi}(H_{\theta}) .
\end{equation}

The frozen LLM generates the report with an autoregressive objective:
\begin{equation}
\mathcal{L}_{\mathrm{gen}}
=
-\sum_{t=1}^{T}
\log p_{\phi}
\left(
y_t \mid y_{<t}, V_{\theta,\psi}
\right) .
\end{equation}

During this stage, the LLM parameters \(\phi\) are frozen. The frozen LLM acts as a semantic decoder, whose pretrained language knowledge and clinical-text priors provide structured supervision. This encourages the visual encoder to produce visual tokens that can be understood and consumed by the LLM for medical report generation.

For training a MLLM for medical visual question answering, the pretrained visual encoder is reused, while the projector and LLM can be replaced by new downstream modules:
\begin{equation}
(x,q)
\rightarrow
E_{\theta}
\rightarrow
P_{\psi^{\ast}}
\rightarrow
F_{\phi^{\ast}}
\rightarrow
a .
\end{equation}
Here, \(q\) is the question, \(a\) is the answer, and the superscript \(\ast\) denotes newly initialized or replaced downstream modules. The projector is replaced because different LLMs use different tokenizers, vocabulary spaces, and hidden embedding dimensions, requiring visual tokens to be mapped into the target LLM's embedding dimension and representation space.

\subsection{Fine-grained evidence preservation}

Report-generation supervision is a language-level signal and may be too coarse to preserve local medical evidence. It can push the visual encoder toward report-level semantics and weaken patch-level visual distinctions, which we refer to as language-induced \textbf{visual collapse}. To reduce this effect, Local Relational Distillation is introduced between the frozen CLIP ViT and the trainable student ViT. The CLIP ViT serves as a stable visual-structure anchor, since it is not updated by the report-generation objective and therefore preserves the original patch-level relational geometry learned during vision-language pretraining

Let \(E_0\) denote the frozen original CLIP ViT teacher and \(E_{\theta}\) denote the trainable student ViT. Given the same image \(x\), we extract their contextualized patch-level visual features after removing the CLS token:
\begin{equation}
H_0
=
E_0^{\mathrm{patch}}(x)
=
[h_0^1,h_0^2,\ldots,h_0^P],
\end{equation}
\begin{equation}
H_{\theta}
=
E_{\theta}^{\mathrm{patch}}(x)
=
[h_{\theta}^1,h_{\theta}^2,\ldots,h_{\theta}^P],
\end{equation}
where \(P\) denotes the number of image patches.

Each patch feature is normalized only for computing patch-to-patch cosine relations:
\begin{equation}
\bar{h}_0^i
=
\frac{h_0^i}{\|h_0^i\|_2},
\qquad
\bar{h}_{\theta}^i
=
\frac{h_{\theta}^i}{\|h_{\theta}^i\|_2}.
\end{equation}

The teacher and student relational matrices are computed as:
\begin{equation}
S_0^{ij}
=
\langle
\bar{h}_0^i,
\bar{h}_0^j
\rangle,
\qquad
S_{\theta}^{ij}
=
\langle
\bar{h}_{\theta}^i,
\bar{h}_{\theta}^j
\rangle .
\end{equation}

The Local Relational Distillation loss is defined as:
\begin{equation}
\mathcal{L}_{\mathrm{LRD}}
=
\frac{1}{P(P-1)}
\sum_{i\ne j}
\left(
S_{\theta}^{ij}
-
\operatorname{sg}(S_0^{ij})
\right)^2 ,
\end{equation}
where \(\operatorname{sg}(\cdot)\) denotes the stop-gradient operation.

This objective does not force the student ViT to copy the teacher's raw patch features directly. Instead, it distills the patch-to-patch relational geometry from a visual teacher. In our implementation, the teacher is the frozen original CLIP ViT; alternatively, it can be replaced by a moving-average teacher updated from the student. Therefore, the trainable ViT can adapt to the frozen LLM generation interface while retaining local visual structure, encouraging fine-grained evidence retention

The final pretraining objective is:
\begin{equation}
\mathcal{L}
=
\mathcal{L}_{\mathrm{gen}}
+
\lambda
\mathcal{L}_{\mathrm{LRD}} .
\end{equation}

\section{Experiment}

\subsection{ViT from MedMLIP improves MLLM over CLIP ViT}
We first investigate whether MedMLIP produces visual representations that
transfer more effectively to downstream medical MLLMs than conventional
contrastive visual pretraining. All visual encoders originate from the same
CLIP ViT initialization and are further pretrained on IU-Xray. We compare three
settings: the original CLIP ViT without additional IU-Xray pretraining,
conventional CLIP-style contrastive pretraining on IU-Xray, and the proposed
MedMLIP pretraining.

Importantly, we evaluate the learned visual encoder under a strict cross-LLM
transfer setting designed to isolate the knowledge encoded in the visual
backbone. Specifically, the MedMLIP encoder pretrained with a frozen
Qwen2.5-0.5B is transferred to a downstream MLLM built upon Qwen2.5-3B, while
the encoder pretrained with Qwen2.5-3B is transferred in the opposite
direction to Qwen2.5-0.5B. These two language models differ not only in
parameter count but also in their internal representation spaces:
Qwen2.5-0.5B uses a hidden and token-embedding dimension of 896, whereas
Qwen2.5-3B uses a dimension of 2,048. They also differ in depth and attention
configuration, employing 24 and 36 Transformer layers, respectively.
Consequently, the visual projection module learned during pretraining is
dimensionally incompatible with the target language model and cannot be
reused.

During transfer, we retain only the pretrained ViT backbone and completely
discard the visual projection module that connected it to the pretraining
LLM. A new projector, whose output dimension matches the embedding space of
the target LLM, is randomly initialized and trained from scratch together
with the downstream MLLM components under the same task-specific training
protocol. Thus, neither the pretraining projector nor the pretraining language
model is retained in the downstream architecture. Any downstream improvement
must therefore be transferred through the visual encoder itself, rather than
arising from reuse of an already aligned projector or continued optimization
with the same LLM. This cross-LLM protocol provides direct
evidence that MedMLIP learns transferable visual representations rather than
an interface specialized to a particular language model.

As shown in Table~\ref{tab:visual_encoder_transfer}, MedMLIP substantially improves
VQA-RAD performance in both transfer directions. When transferred to Qwen3B,
conventional CLIP-style pretraining achieves 67.06\%, whereas MedMLIP with the
full objective reaches 74.20\%, corresponding to an improvement of 7.14
percentage points. Compared with the original CLIP ViT without additional
IU-Xray pretraining, the improvement is 10.71 points. A similar trend is
observed when transferring the Qwen3B-guided visual encoder to Qwen0.5B, where
the full MedMLIP model achieves 61.87\%, compared with 57.14\% for
CLIP-style pretraining and 55.15\% without additional pretraining.

MedMLIP also provides consistent improvements over contrastive IU-Xray
pretraining on SLAKE. With Qwen0.5B as the downstream LLM, the full MedMLIP
encoder reaches 83.14\%, outperforming CLIP-style pretraining by 2.53 points.
With Qwen3B, MedMLIP improves accuracy from 74.14\% to 80.08\%, a gain of
3.94 points. Across the four matched downstream comparisons, the full MedMLIP
encoder improves over conventional contrastive pretraining by between 2.53 and
7.14 percentage points, with an average gain of approximately 4.40 points.

Interestingly, additional IU-Xray pretraining does not universally improve
upon the original CLIP representation. On SLAKE with Qwen3B, the original CLIP
ViT achieves 83.98\%, whereas CLIP-style IU-Xray pretraining reduces the
performance to 74.14\%. MedMLIP recovers part of this degradation and reaches
80.08\%, although it does not surpass the original CLIP baseline in this
setting. One plausible explanation is the limited scale and narrow radiological
coverage of IU-Xray. Continued pretraining on a small chest X-ray dataset may
specialize the encoder at the expense of the broader visual representations
learned by CLIP, particularly when transferring to SLAKE, which contains more
diverse anatomical regions and imaging content. Nevertheless, the improvement
of MedMLIP over CLIP-style pretraining under the same IU-Xray data suggests
that transferability depends not only on the domain data, but also on the
objective used to optimize the visual representation.

Overall, these results support our central hypothesis that visual pretraining
toward an autoregressive LLM interface can produce representations that remain
useful after replacing the language model itself. The cross-LLM transfer
results further suggest that MedMLIP does not merely specialize the visual
encoder to a particular frozen LLM, but instead learns visual features that can
be reused by a different downstream MLLM.

\begin{table}[t]
\centering
\caption{
Transfer performance of different visual encoders on VQA-RAD and SLAKE.
For MedMLIP, the visual encoder is transferred across different LLMs:
the ViT pretrained with Qwen0.5B is evaluated with Qwen3B, and vice versa.
Only the visual encoder is transferred, while the downstream projector and
LLM are replaced.
}
\label{tab:visual_encoder_transfer}

\small
\setlength{\tabcolsep}{3pt}
\renewcommand{\arraystretch}{1.08}

\resizebox{\linewidth}{!}{
\begin{tabular}{lccccc}
\toprule
\multicolumn{1}{c}{\makecell[c]{Visual Encoder /\\Objective}}
& \makecell[c]{Pretraining\\Data}
& \makecell[c]{Guiding\\LLM}
& \makecell[c]{Downstream\\LLM}
& \makecell[c]{VQA-RAD\\ACC}
& \makecell[c]{SLAKE\\ACC} \\
\midrule

CLIP
& --
& --
& Qwen0.5B
& 55.15
& 79.77 \\

BioMedCLIP
& --
& --
& Qwen0.5B
& 61.50
& 72.38 \\

CLIP
& IU-Xray
& --
& Qwen0.5B
& 57.14
& 80.61 \\

MedMLIP ($\mathcal{L}_{\mathrm{gen}}$)
& IU-Xray
& Qwen3B
& Qwen0.5B
& 60.71
& 76.68 \\

MedMLIP ($\mathcal{L}_{\mathrm{gen}}
+\mathcal{L}_{\mathrm{LRD}}$)
& IU-Xray
& Qwen3B
& Qwen0.5B
& 61.87
& 83.14 \\

MedMLIP ($\mathcal{L}_{\mathrm{gen}}
+\mathcal{L}_{\mathrm{LRD}}$)
& Open-PMC-300K
& Qwen3B
& Qwen0.5B
& \textbf{66.66}
& \textbf{86.79} \\

\midrule

CLIP
& --
& --
& Qwen3B
& 63.49
& 83.98 \\

BioMedCLIP
& --
& --
& Qwen3B
& 69.84
& 79.26 \\

CLIP
& IU-Xray
& --
& Qwen3B
& 67.06
& 74.14 \\

MedMLIP ($\mathcal{L}_{\mathrm{gen}}$)
& IU-Xray
& Qwen0.5B
& Qwen3B
& 69.84
& 76.12 \\

MedMLIP ($\mathcal{L}_{\mathrm{gen}}
+\mathcal{L}_{\mathrm{LRD}}$)
& IU-Xray
& Qwen0.5B
& Qwen3B
& \textbf{74.20}
& 80.08 \\

MedMLIP ($\mathcal{L}_{\mathrm{gen}}
+\mathcal{L}_{\mathrm{LRD}}$)
& Open-PMC-300K
& Qwen0.5B
& Qwen3B
& 69.16
& \textbf{84.14} \\

\bottomrule
\end{tabular}
}
\end{table}

\subsection{Scaling MedMLIP and Releasing the Visual Encoder}

We further scale MedMLIP on a larger and more diverse biomedical corpus.
Open-PMC-300K is constructed from the training split of
Open-PMC~\cite{baghbanzadeh2025advancing} through modality-controlled
streaming sampling. We use the \texttt{Sub-caption} field as textual
supervision and retain captions containing between 20 and 1,500 characters.
Samples labeled as non-diagnostic are excluded.

The resulting dataset contains 160,000 radiology images, 110,000 microscopy
images and 30,000 visible-light photographs. These categories correspond to
the R, M and V modality labels in Open-PMC, respectively. In total,
Open-PMC-300K contains 300,000 image--text pairs. Whereas IU-Xray is limited
to chest radiography, Open-PMC-300K includes a wider range of biomedical
imaging modalities and scientific visual content.

We use Open-PMC-300K to pretrain two MedMLIP visual encoders with the same
autoregressive generation and local relational distillation objectives used
in the IU-Xray experiments. One encoder is guided by Qwen0.5B and subsequently
evaluated with Qwen3B. The other is guided by Qwen3B and evaluated with
Qwen0.5B. This follows the same cross-LLM transfer protocol used in our
earlier experiments.

Remember that only the ViT backbone is retained for downstream MLLM training. The
pretraining projector is removed, while a new projector is randomly
initialized and trained from scratch for each target LLM. The downstream
results therefore reflect the transferability of the visual encoder rather to general LLMs
than the reuse of a projector fitted to a specific guiding LLM.

As shown in Table~\ref{tab:visual_encoder_transfer}, the Qwen3B-guided
Open-PMC-300K encoder achieves 66.66\% on VQA-RAD and 86.79\% on SLAKE when
transferred to the Qwen0.5B downstream MLLM. Compared with the corresponding
IU-Xray-pretrained MedMLIP encoder, the improvements are 4.79 percentage
points on VQA-RAD and 3.65 points on SLAKE. The improvement on both datasets
indicates that scaling to a broader corpus produces a more transferable
visual representation in this setting. We therefore select the Qwen3B-guided
ViT as the MedMLIP checkpoint for release.

The Qwen0.5B-guided encoder behaves differently when transferred to Qwen3B.
Replacing IU-Xray with Open-PMC-300K increases SLAKE accuracy from 80.08\% to
84.14\%, but decreases VQA-RAD accuracy from 74.20\% to 69.16\%. A possible
explanation is that Qwen0.5B provides a less expressive autoregressive
supervision signal when the pretraining corpus becomes larger and more
heterogeneous. 

\subsection{Data Quality May Matter More than Data Quantity}
\begin{figure}[t]
    \centering
    \includegraphics[width=1\textwidth, 
                 height=1\textheight,
                 keepaspectratio,     
                 trim=0cm 10cm 0cm 0cm, 
                 clip]{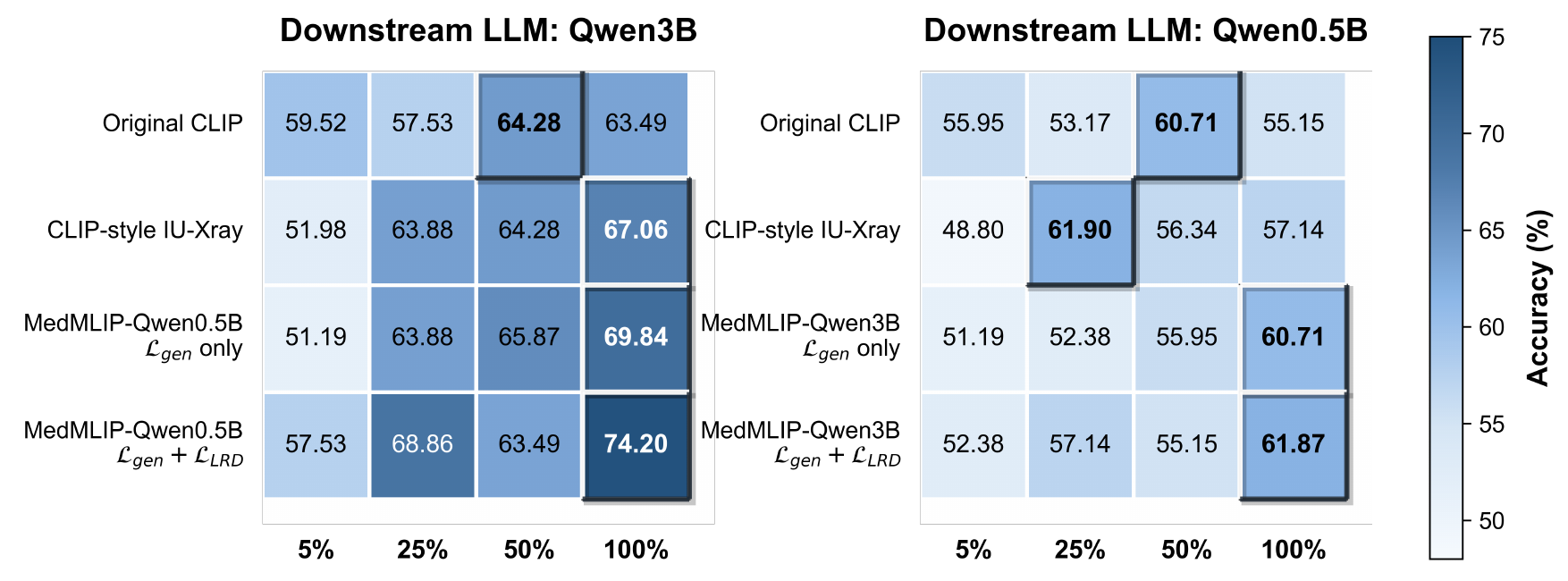}
    \caption{
    Effect of downstream training-set size on VQA-RAD performance.
    We evaluate different visual encoder initialization strategies using
    5\%, 25\%, 50\%, and 100\% of the VQA-RAD training set with Qwen3B
    and Qwen0.5B as downstream language models.
    Highlighted cells indicate the best result within each visual encoder setting.
    Performance is notably non-monotonic with respect to training-set size,
    suggesting that the composition of downstream supervision can have a substantial
    effect beyond the raw number of training examples.
    }
    \label{fig:vqarad_data_composition}
\end{figure}
We further investigate how the amount of downstream supervision affects medical
VQA performance by training with 5\%, 25\%, 50\%, and 100\% of the VQA-RAD
training set. As shown in Fig.~\ref{fig:vqarad_data_composition}, performance
does not increase monotonically with the number of available training samples.
In several settings, substantially smaller subsets outperform larger ones,
despite using the same visual encoder and downstream MLLM architecture.

For example, with CLIP-style IU-Xray pretraining and Qwen0.5B, using only
25\% of the VQA-RAD training set achieves 61.90\% accuracy, compared with
56.34\% using 50\% of the data and 57.14\% using the complete training set.
A similar non-monotonic trend is observed with the original CLIP encoder, where
50\% of the training data reaches 60.71\%, exceeding the 55.15\% obtained
using the full dataset. Since the model architecture and visual initialization
are unchanged within each row, these fluctuations cannot be explained by model
capacity alone.

These results suggest that the effectiveness of downstream supervision depends
not only on its quantity but also on the composition of the selected examples.
A smaller subset may contain a more favorable distribution of informative,
representative, or well-grounded image-question pairs, whereas increasing the
number of training examples may additionally introduce redundant, imbalanced,
or less informative supervision. Therefore, raw training-set size alone is not
a reliable indicator of effective supervision for medical MLLMs.

\subsection{Ablation Study}
We next isolate the contribution of Local Relational Distillation by comparing
MedMLIP trained using only the autoregressive generation objective
$\mathcal{L}_{\mathrm{gen}}$ with the full objective
$\mathcal{L}_{\mathrm{gen}}+\lambda\mathcal{L}_{\mathrm{LRD}}$.
The same cross-LLM transfer protocol is used in both cases, allowing us to
measure whether preserving patch-level relational structure improves the
transferability of the resulting visual encoder.

As shown in Table~\ref{tab:lrd_ablation}, introducing LRD improves downstream
performance in all four evaluated settings. On VQA-RAD, LRD increases
accuracy from 69.84\% to 74.20\% when transferring from Qwen0.5B-guided
pretraining to Qwen3B, corresponding to a gain of 4.36 points. In the opposite
transfer direction, accuracy improves from 60.71\% to 61.87\%.

The effect is particularly pronounced on SLAKE. LRD improves the
Qwen0.5B-to-Qwen3B transfer from 76.12\% to 80.08\%, corresponding to a gain
of 3.96 points. In the Qwen3B-to-Qwen0.5B direction, performance increases
from 76.68\% to 83.14\%, yielding a gain of 6.46 points. Averaged over the two
SLAKE transfer settings, LRD contributes an improvement of 5.21 points.
Across all four experiments, LRD provides an average gain of approximately
3.99 percentage points.

Comparison with the original CLIP encoder further clarifies the role of LRD.
On SLAKE, MedMLIP trained with $\mathcal{L}_{\mathrm{gen}}$ alone performs
below the original CLIP baseline in both transfer settings. With Qwen0.5B as
the downstream LLM, it obtains 76.68\% compared with 79.77\% for CLIP.
With Qwen3B, the corresponding scores are 76.12\% and 83.98\%. Adding LRD
reverses this degradation in the former setting, increasing performance to
83.14\%, and substantially narrows the gap in the latter, reaching 80.08\%.
This comparison suggests that autoregressive objective alone, without visual
preservation may weaken visual structures that remain important for downstream
transfer and therefore does not necessarily produce a more transferable visual
encoder.

\begin{table}[t]
\centering
\caption{
Ablation of Local Relational Distillation (LRD).
Adding $\mathcal{L}_{\mathrm{LRD}}$ consistently improves downstream transfer
over autoregressive generation supervision alone.
}
\label{tab:lrd_ablation}
\small
\setlength{\tabcolsep}{5pt}

\begin{tabular}{lccc ccc}
\toprule
Transfer
& \multicolumn{3}{c}{VQA-RAD}
& \multicolumn{3}{c}{SLAKE} \\
\cmidrule(lr){2-4}
\cmidrule(lr){5-7}
& $\mathcal{L}_{\mathrm{gen}}$
& $+\mathcal{L}_{\mathrm{LRD}}$
& $\Delta$
& $\mathcal{L}_{\mathrm{gen}}$
& $+\mathcal{L}_{\mathrm{LRD}}$
& $\Delta$ \\
\midrule

Qwen0.5B $\rightarrow$ Qwen3B
& 69.84
& \textbf{74.20}
& +4.36
& 76.12
& \textbf{80.08}
& +3.96 \\

Qwen3B $\rightarrow$ Qwen0.5B
& 60.71
& \textbf{61.87}
& +1.16
& 76.68
& \textbf{83.14}
& +6.46 \\

\midrule
Average gain
& --
& --
& \textbf{+2.76}
& --
& --
& \textbf{+5.21} \\

\bottomrule
\end{tabular}
\end{table}

\section{Conclusion}

In this work, we introduced MedMLIP, a visual pretraining framework explicitly
designed for the autoregressive language-model interface used in modern medical
MLLMs. Unlike conventional contrastive pretraining, which aligns visual
representations with a separate text encoder, MedMLIP directly optimizes the
visual encoder according to how its representations will ultimately be consumed
by an LLM. By combining frozen-LLM-guided autoregressive supervision with Local
Relational Distillation, the proposed framework encourages LLM-compatible
semantic representations while preserving the fine-grained and patch-level
visual evidence required for medical understanding.

Experiments on VQA-RAD and SLAKE demonstrate that MedMLIP visual encoders
consistently strengthen downstream medical MLLMs. Importantly, these improvements
remain after replacing both the pretraining LLM and the visual projection
module: only the ViT backbone is transferred, while a new projector is trained
from scratch for a target LLM with a different representation dimension. This
cross-LLM transfer setting indicates that the observed gains arise from improved
visual representations rather than the reuse of a previously learned
visual--language interface. Our ablation results further show that Local
Relational Distillation plays an important role in maintaining transferable
local visual structure during autoregressive pretraining.

Finally, we scale MedMLIP from the radiology-specific IU-Xray dataset to our
more diverse Open-PMC-300K corpus and release the resulting visual encoder as a
reusable initialization for future medical MLLMs. Overall, our findings suggest
that explicitly considering the eventual LLM interface during visual pretraining
is a promising direction for developing more capable, transferable, and
data-efficient medical multimodal systems.

\begin{comment}  %% removed for anonymized MICCAI 2025 submission.
    
    % The following acknowledgement and disclaimer sections should be removed for the double-blind review process.  
    % If and when your paper is accepted, reinsert the acknowledgement and the disclaimer clause in your final camera-ready version.

\begin{credits}
\subsubsection{\ackname} A bold run-in heading in small font size at the end of the paper is
used for general acknowledgments, for example: This study was funded
by X (grant number Y).

\subsubsection{\discintname}
It is now necessary to declare any competing interests or to specifically
state that the authors have no competing interests. Please place the
statement with a bold run-in heading in small font size beneath the
(optional) acknowledgments\footnote{If EquinOCS, our proceedings submission
system, is used, then the disclaimer can be provided directly in the system.},
for example: The authors have no competing interests to declare that are
relevant to the content of this article. Or: Author A has received research
grants from Company W. Author B has received a speaker honorarium from
Company X and owns stock in Company Y. Author C is a member of committee Z.
\end{credits}

\end{comment}

\end{document}